\documentclass[10pt,journal]{IEEEtran}
\ifCLASSOPTIONcompsoc
\else
\fi
\ifCLASSINFOpdf
\else
\fi
\usepackage{graphicx}
\usepackage[cmex10]{amsmath}
\usepackage{amsmath,bm}
\usepackage{multirow}
\usepackage{mathrsfs}
\usepackage{amsmath}
\usepackage{mathtools}
\usepackage{graphicx}
\usepackage[ruled,vlined,linesnumbered,inoutnumbered]{algorithm2e}
\usepackage[table]{xcolor}
\usepackage{color, colortbl}
\definecolor{Gray}{gray}{0.94}
\definecolor{LightCyan}{rgb}{0.88,1,1}
\definecolor{LightYellow}{rgb}{1,1,0.85}
\usepackage{flushend}
\usepackage{underscore}
\usepackage{hyperref}
\usepackage{breakurl}
\usepackage[T1]{fontenc}

\begin{document}
\title{A Study on Stroke Rehabilitation through
Task-Oriented Control of a Haptic Device via
Near-Infrared Spectroscopy-Based BCI}

\author{
Berdakh~Abibullaev,
Jinung~An,
Seung-Hyun Lee,
and~Jeon-Il Moon
\IEEEcompsocitemizethanks{\IEEEcompsocthanksitem 
B. Abibullaev, J.~An, J.I.~Moon,
and S.H. Lee are with the Robotics Research Division. All are with Daegu-Gyeongbuk Institute of Science \& Technology, 
50-1 Sangri, Hyeonpung-Myeon, Dalseong-gun, Daegu, 711-873,
Republic of Korea. 
E-mail: \{berdakho\}@dgist.ac.kr.}
\thanks{}}

\IEEEcompsoctitleabstractindextext{
\begin{abstract}
This paper presents a study in task-oriented approach to
stroke rehabilitation
by controlling a haptic device 
via near-infrared spectroscopy-based brain-computer 
interface (BCI).
The task is to command the haptic device to move
in opposing directions of leftward and rightward movement.
Our study consists of data acquisition, signal preprocessing, and classification. In data acquisition, we conduct experiments 
based on two different mental tasks: one on pure motor imagery, and
another on 
combined motor imagery and action observation. 
The experiments were conducted in both offline and online modes.
In the signal preprocessing, we 
use localization method to eliminate channels that are irrelevant 
to the mental task, as well as perform 
feature extraction for subsequent classification.
We propose multiple support vector machine classifiers
with a majority-voting scheme for improved
classification results.
And lastly, we present 
test results to demonstrate
the efficacy
of our proposed approach 
to possible stroke rehabilitation practice.

\end{abstract}

}
\maketitle

\IEEEdisplaynotcompsoctitleabstractindextext

%
\IEEEpeerreviewmaketitle
\section{Introduction}
Recent rehabilitation methods utilize brain-computer interface (BCI) 
to induce brain plasticity for motor control,
or achieve some degree of patient self-sufficiency by commanding through thinking 
\cite{pfurtscheller2008}.
This approach gained heightened interest
in the research community and 
opened new possibilities for a considerable number of medical applications.
However, there is still a significant work to be performed
before this technology can be fully used in practice, as there
is still no overwhelming evidence of functional recovery on
stroke rehabilitation 
through BCI \cite{gwentrup2011}.

\begin{figure}[t]
\centering
\includegraphics[width=\columnwidth]{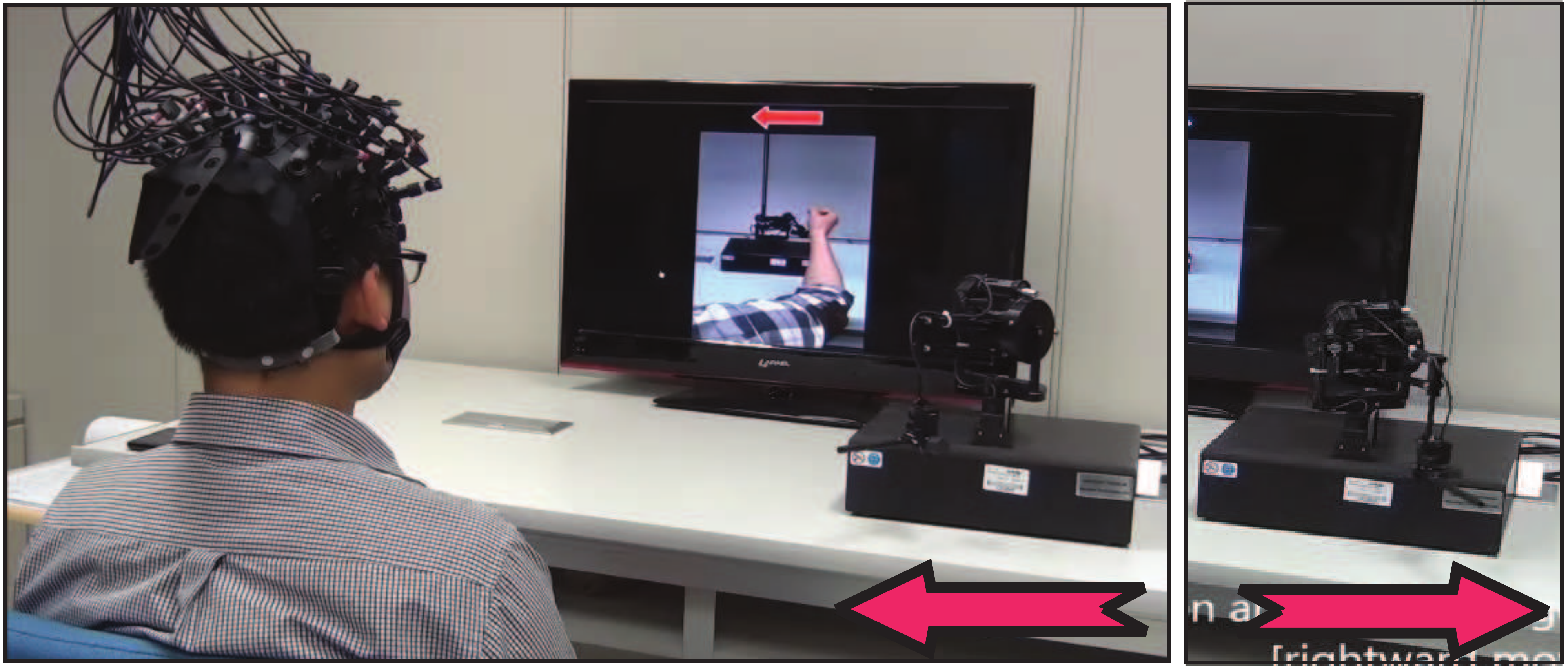}
\caption{A subject commands the haptic device 
to move rightward and leftward through commands
generated by his brain signals and read through
near-infrared spectroscopy-BCI. This experiment
is performed based on both motor imagery 
and action observation. 
A demonstration video of performed
experiments is shown here:
\url{http://youtu.be/bYdJWdPn\_LI}.
}
\label{fig:setup}
\end{figure}

This paper presents a task-oriented training for online 
stoke rehabilitation by controlling a haptic device
through leftward and rightward movement
via 
near-infrared spectroscopy (NIRS)-BCI \cite{coyle2004a},
as shown in Fig.~\ref{fig:setup}.
This emerging modality of non-invasive BCI  
\cite{coffey2010,matthews2008}
measures the cortical hemodynamics and oxygenation status 
through chromophore concentration levels of oxyhemoglobin (oxy-Hb) and deoxyhemoglobin (deoxy-Hb) \cite{zhang2009},\cite{An2013, Abibullaev2013, Abibullaev2014}. 


BCI-based rehabilitation has been the focus of many literature studies which include studies of its different aspects on signal, control, and usage \cite{nijholt2008}; interactive feedback and control strategies\cite{mcrespo2009,krepki2007}; progress of rehabilitation strategies \cite{sitaram2007}; motor imagery to facilitate rehabilitation \cite{dickstein2007}; and implications of BCI to rehabilitation \cite{jerbi2009}.

More recent literature studies looked into virtual reality and its applications to neuroscience research for neurorehabilitation \cite{bohil2011};
BCI in communication \cite{lance2012}, motor
control and neural activities \cite{machado2010}; 
its dependence on
signal acquisition, validation to real-world use, and reliability of function
\cite{shih2012}; and recovery of
hand motor function \cite{mattia2013}.


\begin{figure*}[t]
\centering
\includegraphics[width=0.72\textwidth]{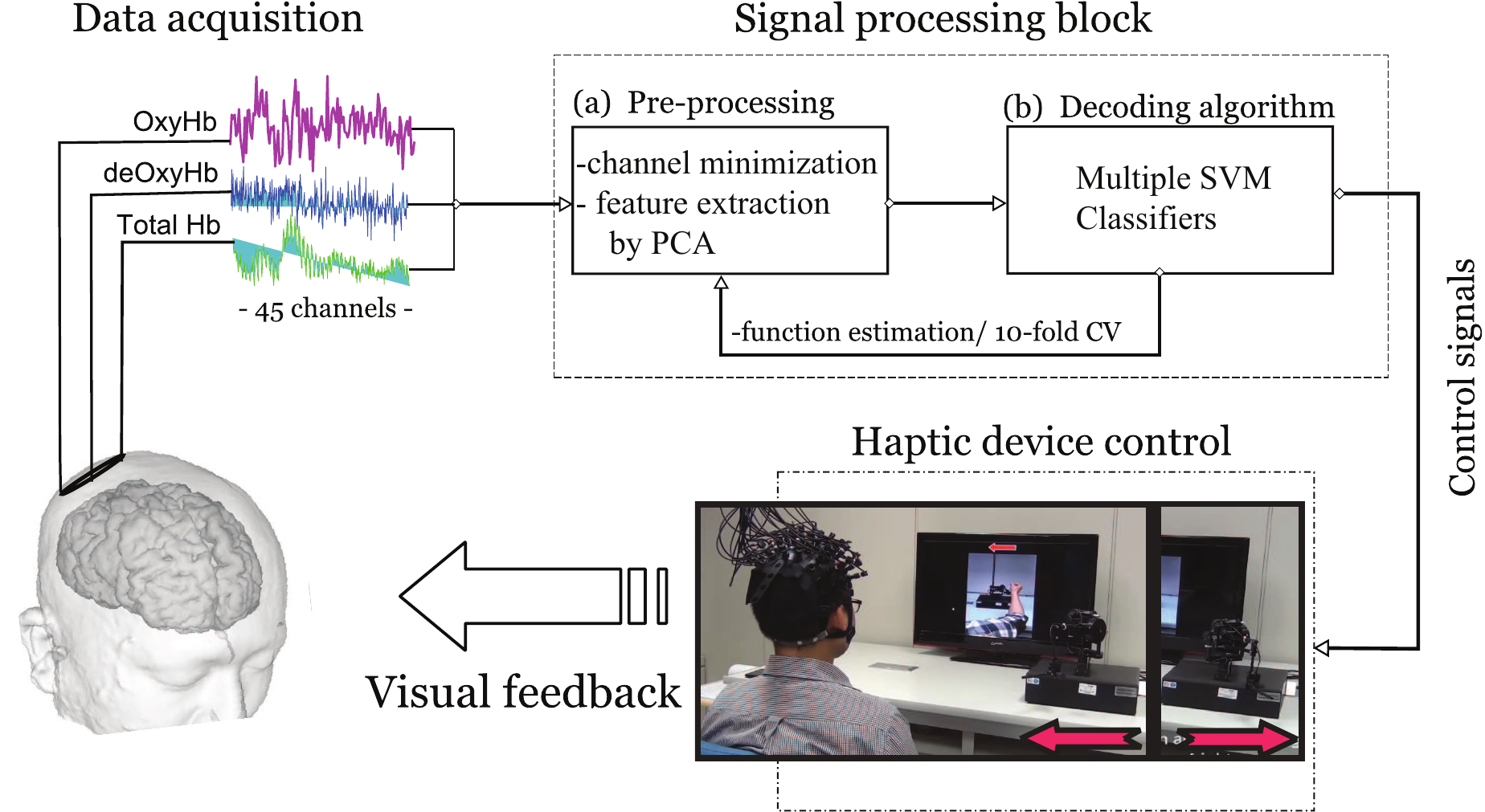}
\caption{Flowchart of the task-oriented control
of a haptic device via NIRS-BCI. 
Concentrations of 
oxy-Hb and deoxy-Hb are read from 45-channel of NIRS.
Input signals are pre-processed by identifying the
more significant channels, and by performing feature extraction through principal component analysis (PCA). Classification
is performed via multiple support vector machines (SVMs), whose output
is also used for tenfold cross validation
(CV).
The decoded outputs are then used to control movements of an external haptic device
in either leftward or rightward motion. 
Success or failure of the task required is determined
through the visual feedback of the haptic motion.}
\label{fig:flowchart}
\end{figure*}


A considerable number of experiments related to BCI-based
rehabilitation have been conducted. These
include
minimal training and mental stress to patient \cite{bai2010},
rehabilitative intervention for hand plegia \cite{buch2008},
control of 9-degrees-of-freedom (DOF) wheelchair-mounted robotic arm \cite{palankar2008}, and
virtual environment to facilitate neuroplasticity \cite{merians2009}.
More recent 
experiments
include 
detection of movement intention \cite{niazi2011},
exoskeleton to control fingers with feedback \cite{rmurguialday2012},
removing artifact in motor imagery \cite{murguialday2010},
calibrating imagery through passive movement \cite{ang2011},
studying motor learning after stroke \cite{meyer2012}, and
test of feasibility of single-trial, individually-tuned classifiers
\cite{zimmermann2013}.

Despite the above efforts, only a few BCI-based rehabilitation
studies have included a haptic device 
in their approach.
Interestingly, there are empirical evidences that tactile sensing through
haptic feedback \cite{grodriguez2011}
and vibro-tactile stimulus \cite{chatterjee2007} 
show improved
rehabilitation results. 
Our work aims to contribute to the same efforts of including
haptic device to BCI-based rehabilitation, and is implemented
by using both motor imagery and 
combined motor imagery-action observation methods.
To drive the haptic device, signals are pre-processed
and most significant channels are identified, then the 
output signals are classified to move the haptic device
to a desired direction.
Visual feedback determines the success or failure of the
desired action based on the subject's brain signal command.
Our experimental setup will be described and 
results of training the classifier will be shown.
Online and offline test results will be presented which determines
the efficacy of our proposed method for stroke rehabilitation.

This work proceeds as follows. 
Section~\ref{sec:matmethods} presents how our classifiers are
trained and optimized through offline supervised learning.
Once these classifiers are optimized, we test them with offline
and online data sets. 
The offline test data results 
are shown in Section~\ref{sec:resultsoffline},
while the more challenging case of testing via online
data streaming is shown in Section~\ref{sec:resultsonline}.
And lastly, Section~\ref{sec:discussion} shows the discussion and 
comparison between our results and the previously published results
in BCI-based rehabilitation.

\section{Materials and Methods}
\label{sec:matmethods}

Offline supervised learning is used to train and optimize
our classifiers. First, raw signals are pre-processed 
using feature extraction
through
principal component analysis (PCA).
This reduces the noise from the raw signal that was
read through NIRS-BCI.
Furthermore, more significant channels of NIRS-BCI are identified
through recursive channel elimination (RCE). This eliminates
the non-task-relevant channels, which can be another source of
signal noise. From the processed signals that were
read through task-relevant channels, classification is 
performed based on the actions commanded by the subject. 
Our classification uses multiple support vector machines (SVMs),
where a majority voting mechanism is then used to further refine the classification
process.
Output from SVMs is used for tenfold cross validation
(CV) in the signal pre-processing stage.
Test data from both offline and online 
data sets verifies the efficacy of the trained classifiers. 
The flowchart of the entire experimental 
process is shown in 
Fig.~\ref{fig:flowchart}.

\begin{figure*}[t]
\centering
\includegraphics[width=0.7\textwidth]{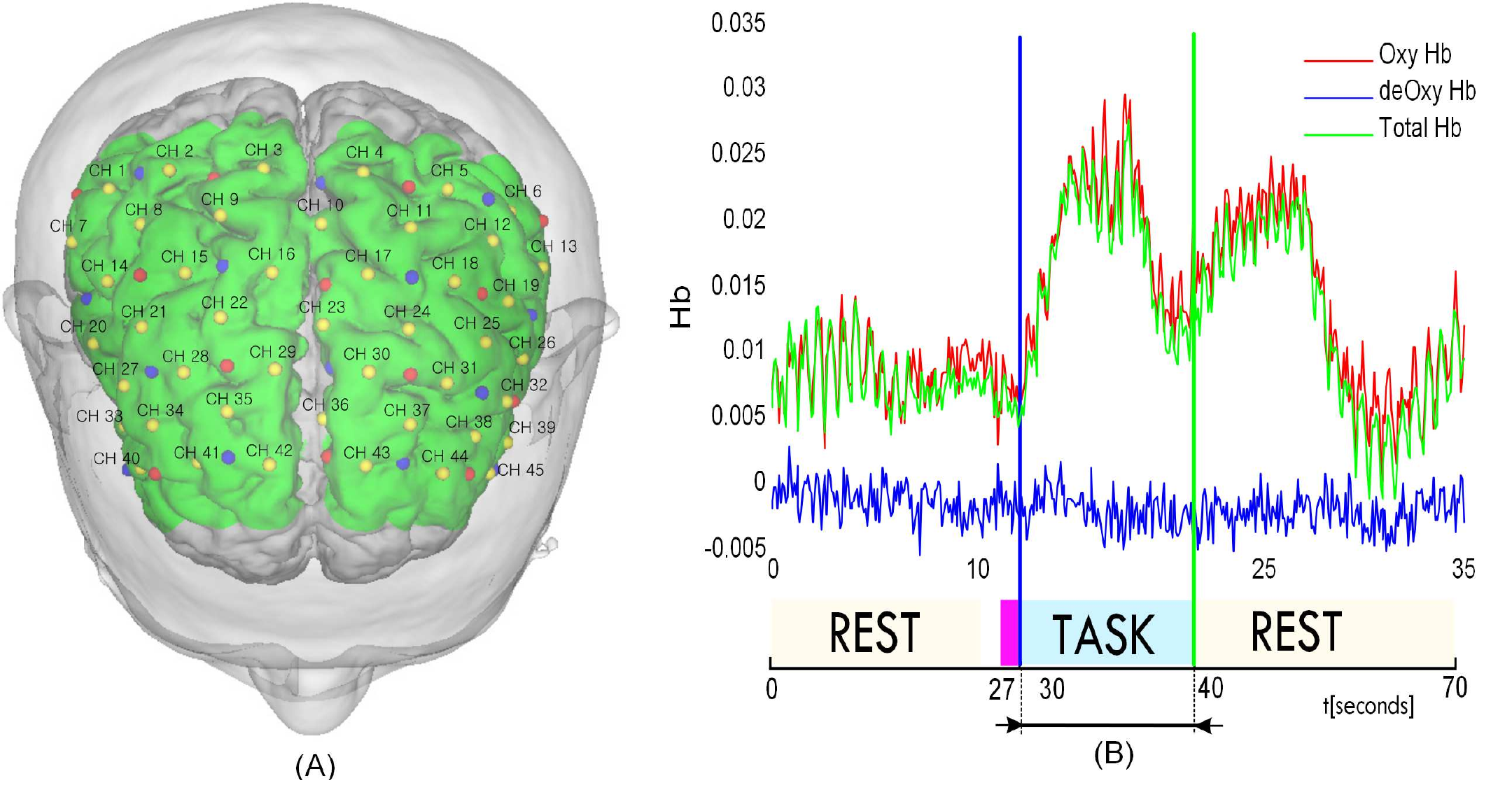}
\caption{(A) The locations of NIRS emitter-receiver optodes with 30-mm interoptode distance. The red circles represent emitters and blue circles represent receivers. The yellow circles represent the locations of the 45 channels recorded. (B) The timing of a single experimental trial
of data acquisition, shown with its corresponding
oxy-Hb, deoxy-Hb, and total Hb concentrations at each
stage of the task.}
\label{fig:session}
\end{figure*}

\subsection{Data Acquisition}

Experimental subjects consist of  
seven healthy, right-handed males ages $28\pm 4$ years. 
All study participants gave informed consent. The ethical approval of the research was granted by the research ethics committees of the Daegu-Gyeongbuk Institute of Science and Technology.

The NIRS-BCI used in our work has 45-channel
optical brain-function imaging system for data acquisition (FOIRE-3000, Shimadzu Co. Ltd., Japan). 
It uses safe near-infrared light to assess the oxy-Hb and deoxy-Hb concentrations of the brain at wavelengths of 780 nm, 805 nm, and 830 nm. This study uses
concentration levels of oxy-Hb for analysis and classification, which are found to be more correlated with the regional cerebral blood flow (rCBF) than deoxy-Hb and total-Hb \cite{gratton2005}.
An increase in rCBF reflects an increase in neural activity \cite{jueptner1995}. 

We placed the optical fiber probes on the frontoparietal regions of the brain cortex to cover an area of $21\times 12$ cm as shown in Fig.~\ref{fig:session}A. The subjects performed three types of mental tasks denoted by  $\{t_{right}\}$, $\{t_{left}\}$ and $\{t_{rest}\}$ as follows:

\begin{itemize}
\item $\{t_{right}\}$ - subjects repetitively performed an imaginary rightward movement of the haptic device,
\item $\{t_{left}\}$ -  subjects repetitively performed an imaginary leftward movement of the haptic device, and
\item $\{t_{rest}\}$ - subjects rest and perform no actual task. 
\end{itemize} 

The signals during rest were used as the baseline in a classification process.
Each subject performed five-session mental tasks
for a total of 35 sessions for all subjects. 
We split every session into three blocks $[Rest \rightarrow Task\rightarrow Rest]$ as shown in Fig.~\ref{fig:session}B. 
In the same figure, the corresponding levels of
oxy-Hb, deoxy-Hb, and total Hb are also shown 
during one experimental session
of MI task. 

\subsection{Types of Experiments}
This work uses two types of tasks to control the
haptic device: 1.) motor imagery (MI) task, and
2.) a combined action observation (AO) and MI tasks.
The latter type of task is also referred to as AOMI task.

In MI tasks, subjects merely imagine 
the 
task 
without an external cue.
In AOMI tasks, the subject performs an 
AO task followed by an
MI task. The AO task
consists of watching a video that shows the movements of a subject’s forearm in the intended direction.
Our motivation for the AOMI task experiment
is based on earlier studies related to the putative human mirror neuron system that describe how predictions and interpretations of the actions of others were exploited for BCI systems \cite{jarvelainen2004} \cite{tkach2007}.
We want to investigate whether the combined AOMI task provides higher BCI classification rates than a pure MI task.

\subsection{Signal Pre-processing}

We consider two significant factors that affect the accuracy of a BCI system: 1.) background noise, and 2.) task-irrelevant channels.

The noise interference in hemodynamic signals may arise from instrumental, experimental, or physiological sources. 
Particularly, physiological noise often overlap in frequency with the expected neural signals \cite{coyle2004a}. 
In this study, we employ PCA for noise reduction and 
feature extraction,
which has 
shown to be more reliable in eliminating background noises in NIRS signals \cite{virtanen2009}.
Other noise-reduction methods use Weiner filtering \cite{izzetoglu2005}, wavelets \cite{jang2009,abibullaev2012}, and adaptive filtering \cite{zhang2009}.

Selecting task-relevant channels may yield the
required accuracy with greater convenience \cite{lal2004,schroeter2004}. 
Unfortunately, optimal channel selection is not a trivial task, in particular for NIRS-BCI, when extracting neurophysiologic knowledge corresponding to a specific mental task. The next section describes our BCI channel selection strategy in more detail.

%
%
%

\subsection{Principal Component Analysis}  
The neural datasets are denoted by $X\in R^{l\times m}$,
such that $l$ is the number of channels and
$m$ is the number of samples.
We denote a single trial dataset as a matrix  
$X\in R^{l\times m}$  
that has its rows $\{{x_1},{x_2},...,{x_l}\}^T$ composed of the channel observations with $m$ features or dimensions. Our goal is to find a new data matrix $X\in R^{l\times k}$ where $k<m$.

We employ a PCA method for this purpose. It is based on projecting signal features ${x}\in R^{m}$ onto a subspace defined by a set of orthonormal vectors $u\in R^m$ that maximize the data variance $E$,
\begin{eqnarray}
\label{eqn:maxE}
&&\mbox{maximize}\;\;\;  E = u^T X^T Xu\\
&&\mbox{subject to}\;\;\; \|u\| = 1\nonumber
\end{eqnarray}

Solving the optimization in Eq.~\ref{eqn:maxE} by the Lagrangian method yields the eigenvalue equation $X^T Xu = \lambda u$. 
It follows that to maximize the variance,
the chosen $u$ 
must be the eigenvector of $X^T X$ corresponding to the largest eigenvalue. In order to compute $k$ directions, we must find eigenvectors $u_1,...,u_k$ corresponding to the $k$ largest eigenvalues given $\lambda_1,...,\lambda_k,$
such that
$\lambda_1\geq\lambda_2...\geq\lambda_k$.
Algorithm~\ref{alg:pca} shows our method to find the PCA projection directions.
The resulting features extracted by PCA are $Xu_1,...,Xu_k$. 

\begin{algorithm}
\KwIn{Data matrix $X^{l\times m}$, dimension $k$}
Process: $X_1 =X$\;
\ForEach{$i = 1,...,k$}{
Select $u_j$ as the first eigenvector of $X_{i}^T X_j$\;
$X_{j+1} = X_{j}\big(I-\frac{u_{j}u_{j}^T }{u_{j}^T u_{j}}\big)$}
\KwOut{
Projection directions $u$ and features $Xu$}
\caption{PCA Pre-procesing}
\label{alg:pca}
\end{algorithm}

\subsection{Recursive Channel Elimination}
We employ a recursive channel elimination (RCE) algorithm \cite{lal2004},\cite{schroder2005} for 
identifying
the recording positions most relevant to cognitive tasks. The method is based on recursive feature elimination \cite{guyon2002}, which is an iterative, embedded, greedy backward method of feature selection. The best channels are determined by training several SVMs and exploring their marginal characteristics. Algorithm~\ref{alg:rce} describes our method for channel selection.
The algorithm can can be summarized as follows:

\begin{itemize}
\item make ten disjoint training datasets (Line 2) for tenfold 
CV, 
\item train a linear SVM (Line 9) and estimate the generalization error (Line 10)
for each fold,
\item estimate the rank of the channels based on a margin ranking criterion
(Lines 11-14), and
\item eliminate channels with the lowest ranking score criterion (Line 15). 
\end{itemize}

We repeat the procedure until the required number of channels is retained throughout all ten datasets. We define a threshold value for the number of channels that potentially need to be retained. This is done by trial and error on the basis of the test error rate at Line 10. We tried several channel combinations and decided to select only 20 of the 45 channels for subsequent classification.


\begin{algorithm}
\KwIn{$\{x_{i},y_{i}\}_{i=1}^{l}$, $x_i\in X$, $y_i\in\{\pm1\}$, 
training set with $l$ channels related to either $\{t_{right}\}$ or $\{t_{left}\}$ tasks.}
Perform tenfold, divide the training set (of size) $m$ into $p$ disjoint sets $S_j,...,S_p$ of equal size $p/m$\; 
\ForEach{$S_j$}{
Initialize: $Ranked = [\emptyset]$\; 
Surviving channels  $Ch_{j} = [1,2,...,l]$ 

\While{Surviving channel is not empty}{
\ForEach{channel in $Ch_{j} = [1,2,...,l]$}{
Temporarily remove channel $j$ in $Ch_{(j=1,...,l)}$\;
Train a linear SVM with the remaining channels of $S/S_j$ and estimate $|w|$ (from Eq.~\ref{eqn:minalpha}, Eq.~\ref{eqn:fz})\;
Test it on $S_j\mapsto \{\pm1\}$\;
Compute the ranking score: $R_{j} =\frac{1}{|Ch_{j}|}\sum_{l\in Ch_{j}}|w_{l}| $\;
}
Locate channels with smallest ranking criterion: 
$RankChan = argmin\{R_{j}\}$\;
Update channel rank: $Ranked = [RankChan,Ranked]$\;
Eliminate the channel with smallest $R_{j}$ score\;}
}
\KwOut{Extracted channel list: $Ranked$}
\caption{Recursive Channel Elimination}
\label{alg:rce}
\end{algorithm}

\subsection{Classification}
Given set of pre-processed training dataset  $X:=\{x_1,...,x_m\}$ with corresponding labels $Y:=\{y_1,...,y_m\}$,
where $y_i\in\{\pm1\}$ for $i=1,\ldots,m$, 
our next goal is to estimate a function $f:X\rightarrow \{\pm 1\}$ to predict whether a new signal observation  $z\in X^{*}$ will belong to class $+1$ or $-1$. 
We defined the classes for the mental tasks $[\{t_{right},+1\},\{t_{rest},,-1\}]$ as patterns related to rightward movement ($y=+1$) and the baseline ($y=-1$). Similarly, we define $[\{t_{left},+1\},\{t_{rest},-1\}]$ as patterns related to leftward movement and the baseline ($y=-1$). We estimate a set of SVM functions for classification with a soft margin loss function $L(x,y,f(x)) = \mbox{max}(0,1-yf(x))$. 

The solution of SVM is based on the following optimization \cite{scholkopf2002}:,

\begin{eqnarray}
\label{eqn:minalpha}
&& \operatorname*{min}_{\alpha\in R,b\in R} \bigg\{\frac{1}{C}\sum_{i=1}^{m}\xi_{i}+\sum_{i,j=1}^{m}\alpha_{i}y_{i}K(x_i,z_j)\alpha_{j}y_{j}\bigg\} \nonumber\\
&& \mbox{subject to}\;\;\;  y_j\bigg(\sum_{i=1}^{m}\alpha_{i}y_{i}K(x_{i},z_{j})+b\bigg)\geq 1-\xi_{i} \nonumber\\
&&\;\;\;\;\;\;\;\;\;\;\;\;\;\;\;\;\;\;\; \xi_i\geq 0, \forall i = 1,2,...,n 
\end{eqnarray}

\noindent where $\alpha = (\alpha_{1},...,\alpha_{m})$ are Lagrange multipliers, the $\xi_{i}$ are slack variables and a user defined regularization parameter $C>0$. The corresponding decision function is given by,

\begin{equation}
\label{eqn:fz}
f(z) = \mbox{sign}\bigg[\big(\sum_{i=1}^{m}\alpha_{i}y_{i}K(x_{i},z)\big)+ b\bigg].
\end{equation}

From Eq.\ref{eqn:minalpha} and Eq.~\ref{eqn:fz}, the $K(x,z)$ is a reproducing kernel function which gives rise to a Gram matrix $K_{i,j}:=(x_i,z_j)$, $K\in R^{m\times m}$ [\textbf{20}]. This matrix contains all the information available in order to perform data analysis and modeling of SVM algorithm. 
Note that we use this formulation of the SVM for two different purposes. First, we use SVMs for the recursive channel elimination method in Algorithm~\ref{alg:rce}. Second, SVMs constitute the bases functions of multiple classifiers which we use for decoding of signal features related to MI and AOMI tasks.  

\begin{algorithm}
\textbf{Define}:  $\mathcal{E}_{1} = \{f_{1},...,f_{6}\}$ and $\mathcal{E}_{2}=\{f_{1}^{*},...,f_{6}^{*}\}$ \; 
\KwIn{$\{x_{i},y_{i}\}_{i=1}^{m}$, $x_i\in X$, $y_i\in\{\pm1\}$, 
training set related to either $\{t_{right}\}$ or $\{t_{left}\}$ mental tasks.}
\ForEach{$f_{i}\in\mathcal{E}_{1}$ or $f_{j}^{*}\in\mathcal{E}_{2}$, $i,j = 1,...,6$ }{
Perform tenfold CV and a search on optimal $C$\;
Divide the training set (of size) $m$ into $p$ disjoint sets $S_j,...,S_p$ of equal size $p/m$\; 
\ForEach{$S_j$}{
Train a $f_{i}(x)$ on $S/S_j$\; 
Test it on $S_j\mapsto AUC(j)$\;
\KwOut{Optimized classifier model : $f(\cdot,\alpha,b)$}
}}
\KwOut{
Set of optimized $\mathcal{E}_{1} = \{f_{1}(\cdot,\alpha_{1}),...,f_{6}(\cdot,\alpha_{6})\}$               $\mathcal{E}_{2}=\{f_{1}^{*}(\cdot,\alpha_{1}),...,f_{6}^{*}(\cdot,\alpha_{6})\}$.
}
\KwIn{$\{z_{i}\}_{i=1}^{n}$, $z_i\in X$ unseen test patterns related to either $\{t_{right}\}$ or $\{t_{left}\}$ mental tasks.}
\ForEach{$f_{i}(\cdot,\alpha_{i})\in\mathcal{E}_{1}$ or $f_{j}^{*}(\cdot,\alpha_{j})\in\mathcal{E}_{2}$, $i,j = 1,...,6$}{
\textbf{Evaluate $\mathcal{E}$}: $f_{i}(x_{j},\alpha_{i})\rightarrow y_{N,l}$, where $y_{j,i}\{\pm1\}$ are the columns of Table 1, $i= 1,...,l$ classifiers and $j = 1,...,n$ test patterns\;
(1)-\textit{Majority vote}: define $k$-of-$l$ majority voting classifiers as defined in Eq.~\ref{eqn:FzU}.\;
(2)-\textit{Output}: compute the final AUC value for majority classifiers $\mathcal{E}\rightarrow AUC$\;}
\KwOut{$\mathcal{E}_{1}\rightarrow AUC_{1}$ and $\mathcal{E}_{2}\rightarrow AUC_{2}$
}
\caption{Multiple SVM Training and Testing}
\label{alg:msvm}
\end{algorithm}
\begin{table}[h!b!p!]
\caption{Structure of multiple outputs}
\centering 
\begin{tabular}{c|c c c c c}
\hline
\\[-4pt]
 ~      & $f_1$ & $\cdots$ & $f_i$ & $\cdots$ & $f_n$ \\ 
\hline\vspace{4pt} 
$x_1$  & $y_{1,1}$ & $\cdots$ &  $y_{1,i}$ & $\cdots$ &  $y_{1,n}$ \\ 
$\vdots$        & $\vdots$ & ~ & $\vdots$ & ~ & $\vdots$ \\ 
$x_j$  & $y_{j,1}$ & $\dots$ &  $y_{j,i}$ & $\dots$ &  $y_{j,n}$ \\ 
$\vdots$        & $\vdots$ & ~ & $\vdots$ & ~ & $\vdots$ \\ 
$x_m$ & $y_{m,1}$ & $\dots$ &  $y_{m,i}$ & $\dots$ &  $y_{m,n}$ \\ 
\end{tabular} 
\end{table}

\subsection{Multiple SVM Classifiers}
Instead of training a single classifier, we study train multiple SVMs with the purpose of further improving the overall BCI accuracy. We consider a multiple $n$ - classifier functions $\{f_1,f_2,...,f_n\}$ and a data set $\{(x_i,y_i)^{m}_{i=1}\},x_i\in X,\; y\in Y$. Each classifier is trained independently to predict  $f_{i=1}^{n}:x\rightarrow \{\pm 1\}^{n}$. The outputs from all classifier functions are then defined as an $m$-dimensional binary vector $y = [y_{1,i},...,y_{m,i}]$, such that $y_{j,i} = 1$ if $f_i$ recognizes $x_j$ and 0 otherwise for $i = 1,...,n$. 

Table 1 shows that the number of correct assignments is  $N_1(f_i) = \sum_{j=1}^{m}y_{j,i}$ and the number of mistakes is $N_0(f_i) = m - \sum_{j=1}^{m}y_{j,i}$. In order to make the final decision from the set of functions $\{f_i,...,f_n\}$, we define the following majority voting rule:
\begin{equation}
\label{eqn:FzU}
  F(z) = \left\{
  \begin{array}{l l}
    +1  & \quad \textrm{if\;\;\;  $\sum_{i}^{n}f_{i}(z)\geq k$}\\
    -1  & \quad \textrm{else $\sum_{i}^{l}f_{i}(z)\leq n-k$}\\
     U  & \quad \textrm{Otherwise}\\
  \end{array} \right.
\end{equation}

\noindent where $k<n$ and $i = 1,...,k$ making similar predictions defined by the $k-\mbox{of}-n$  majority  classifier for $k\geq\frac{n}{2}$. In this case, $U$ represents the unknown outputs or failure in predicting both outputs. Thus, we have three possible outcomes from all classifiers $F:X\rightarrow \{+1,-1,U\}$. 

Algorithm\ref{alg:msvm} gives the details of the proposed multiple SVM classifier training and validation sequentially. This consists of two main phases, namely, the training phase and the testing phase. In both phases, we train and test two different group of multiple classifiers  $\mathcal{E}_1$ and $\mathcal{E}_2$. 

The group $\mathcal{E}_1$ is trained by taking the examples from rightward task $\{t_{right}\}$ as positive and the examples from the rest task $\{t_{rest}\}$ as negative. Likewise, the group $\mathcal{E}_2$ is trained by taking the examples from the leftward task $\{t_{left}\}$ as positive and the examples from the rest task $\{t_{rest}\}$ as negative. Each group consists of six base SVM functions with linear kernels. 

In the training phase, each individual base SVM function is trained separately using the same input data from the tenfold CV (Algorithm~\ref{alg:msvm}, Lines 1-9). 
  
During the testing phase, unseen examples are applied to all base functions simultaneously in real time. Further, a collective decision is obtained on the basis of the majority voting scheme using Eq. 4 (Algorithm~\ref{alg:msvm}, Lines 12-16). In other words, once each of the six base classifiers has cast its vote, the majority voting strategy assigns the test patterns to the class with the largest number of votes and outputs are provided as the final prediction.


Then, the final decision on which direction to move the haptic device with the output control command is based on the area under the receiver operating characteristic (ROC) curve. This area under the ROC curve is also referred to as AUC. 
The AUC is a comparatively robust measure that is insensitive to class distributions and misclassification costs \cite{bradley2009}. For instance, AUC $=1$ indicates perfect classification, whereas AUC $=0.5$ indicates that the result from the classifier is no better than a random guess.
In our case, an AUC $>0.70$ moves the haptic device to the desired direction.

\begin{figure*}[htbp]
\centering
\includegraphics[width=0.9\textwidth]{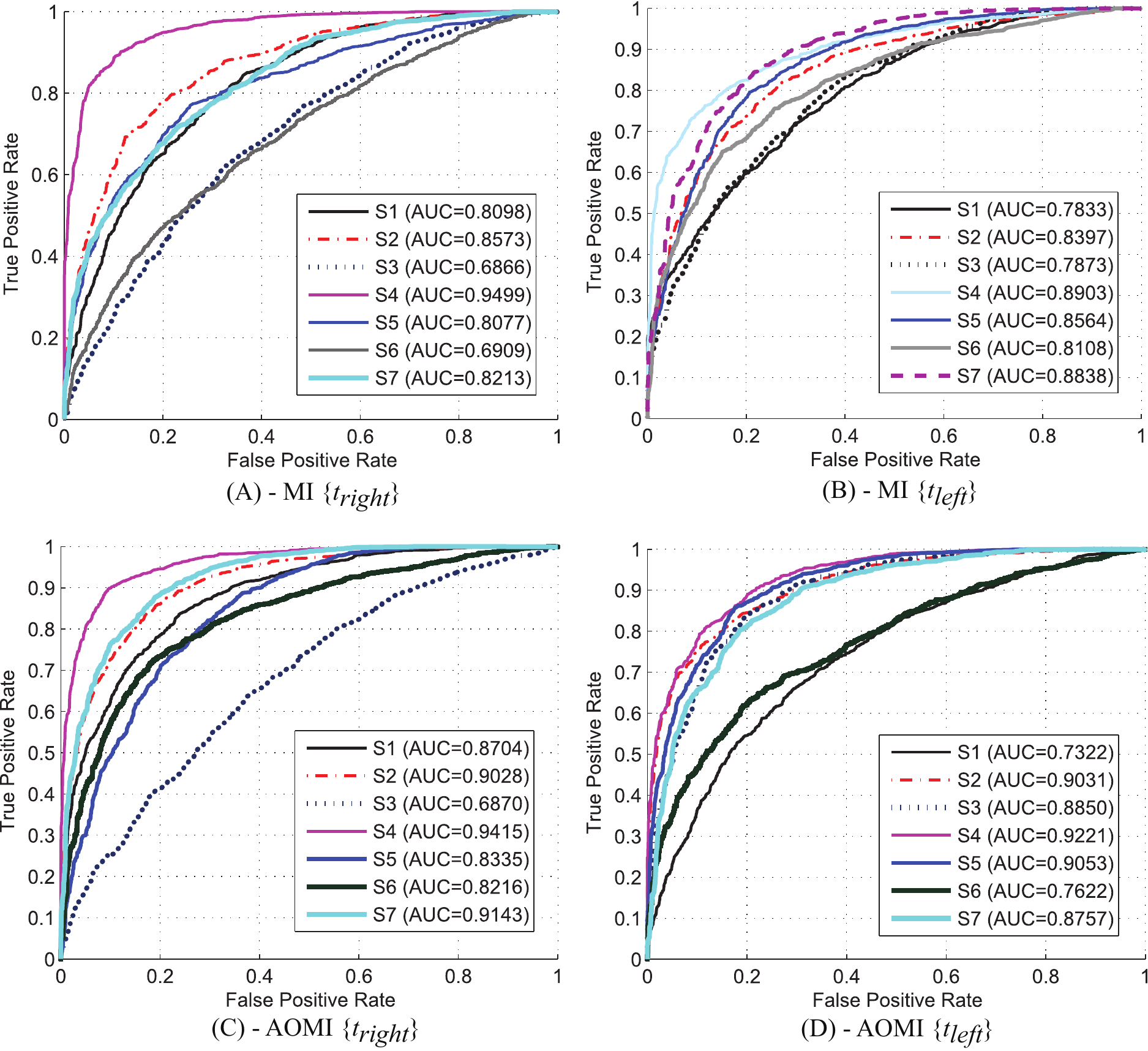}
\caption{Offline classification results from seven subjects for random test patterns. Plots (A) and (B) represent decoding of motor imagery (MI) tasks: (A) for $\{t_{right}\}$ task, and (B) for $\{t_{left}\}$ task. Plots (C) and (D) represent decoding of action observation-motor imagery (AOMI) tasks: (C) for $\{t_{right}\}$, and (D) for $\{t_{left}\}$. The value of AUC $>0.70$ is acceptable.}
\label{fig:offlineresults}
\end{figure*}

\section{Offline Test Results and Analysis} 
\label{sec:resultsoffline}

For each of the seven subjects, we trained subject-specific multiple SVM
classifiers (Algorithm~\ref{alg:msvm}) with an input dataset consisting of 20 channels selected using the RCE algorithm. 
The relevant channel locations varied among subjects and sessions, and were updated every time a subject performs mental tasks. 
The search for an optimal penalty parameter was conducted
to obtain the best CV performance. 
Then we gathered offline data set to test the performance of our 
resulting classifier. The resulting UAC values are shown in Fig.~\ref{fig:offlineresults} in moving the haptic device in rightward and leftward
direction, in both MI and AOMI task commands.
We assign AUC $>0.70$ to be acceptable, such that the haptic device is moved
to the desired direction.

For the sake of discussion, we introduce the following three classifier performance regions:

\begin{itemize}
\item $\Theta_{best}\;\;\; := (0.80,...,1]$,\;\;\;\;\; if AUC $\geq 0.80$ ,
\item $\Theta_{accept} := (0.70,...,0.80]$, if $0.70 < \mbox{AUC} \leq 0.80$ ,
\item $\Theta_{worst\;}  := (0.60,...,0.70]$, if $0.60 < \mbox{AUC} \leq 0.70$ .\\
\end{itemize}

The haptic system is the PHANTOM Premium 1.0 haptic device (19.5 cm $\times$ 27 cm $\times$ 37.5 cm workspace, two active degrees of freedom). Real-time neural data were acquired through a LabVIEW - NIRS interface, wherein the proposed algorithms were implemented.

Let us consider the
decoding results of
MI task commands in
Figs.~\ref{fig:offlineresults}A (rightward movement) and 
\ref{fig:offlineresults}B (leftward movement).
In particular, for the rightward movement
two subjects showed superior results: $S4$(AUC=0.9499) and $S2$(AUC=0.8573).
In addition, three more subjects showed satisfactory results: 
$S1$(AUC=0.8098), $S5$(AUC=0.8077), and $S7$(AUC=0.8213). 
However, we noted inferior classifier performance for the remaining two subjects: $S3$ (AUC=0.6866) and $S6$ (AUC=0.6909).
For the leftward movement, 
all subject showed satisfactory results, with three subjects 
$S4$, $S5$, and $S7$ showing superior results.

Inconsistency of results in the offline mode can be attributed to
BCI intersession variability \cite{gerven2009}. 
This problem of dramatic variability arises in neural signal measurements obtained during different recording sessions, even when the same subject is used. In addition, many other factors may affect the characteristics of neural signal measurements, resulting in variations. Such factors include the subject's condition, mood, fatigue, and drowsiness or even the subject's level of attention to a particular mental task \cite{blankertz2008}.

We subsequently investigated 
AOMI task commands. The plots 
of the decoding signal patterns
are shown in 
Figs.~\ref{fig:offlineresults}C for the rightward task command, and
Figs.~\ref{fig:offlineresults}D for the leftward task command.
Significant improvements in AUC values with AOMI tasks were noticeable.
For instance, superior results were observed for four subjects
($S1$, $S2$, $S4$, and $S7$) in the rightward command task, with only
one subject ($S3$) that consistently remained with unsatisfactory results
in the rightward command task. In addition, all subjects showed improved
performance in AOMI compared to MI, with only one subject, $S4$,
degrades by AUC=0.0084, but still showed superior results at AUC=0.9415.
One subject, $S6$, the previously showed unsatisfactory results with MI
now has a much better performance with an AUC=0.8216.
For the leftward command task, all the subjects showed satisfactory
results with four subjects ($S2$,$S3$,$S4$,$S5$) showing superior results. 
Degraded results showed for subjects $S1$, $S6$, and $S7$, but remained
within satisfactory results.


In summary, the offline results showed that our proposed classification
method in reading brain signal commands to move the haptic device leftward and
rightward, has successfully achieved the desired motion at 
25 out of 28 test cases
with only three that 
showed unsatisfactory
results.
Furthermore, we observed that signal patterns using AOMI task commands produced better classification results than those 
using pure MI task commands. This offline analysis may seem uninformative so far. However, a more important aspect of our research is stable performance of the 
derived classification models during the real-time BCI experiments. 
In the next section we report the online results.

\begin{table*}[ht]
\caption{Experiment 1: The online performance of classifiers in decoding signals corresponding to pure motor imagery (MI) task commands of 
$\{t_{right}\}$, $\{t_{left}\}$, and $\{t_{rest}\}$.} 
\centering 
\begin{tabular}{l | c c c|c c c| c c c} 
\hline\hline\\[-5pt] 
Subjects &~ &  Session 1 & ~ & ~ & Session 2 & ~ & ~ &  Session 3  & ~\\ 
\\[-2ex]
\rowcolor{LightCyan}
~ & $\{t_{right}\}$ & $\{t_{left}\}$ & $\{t_{rest}\}$ & $\{t_{right}\}$ & $\{t_{left}\}$ & $\{t_{rest}\}$ & $\{t_{right}\}$ & $\{t_{left}\}$ & $\{t_{rest}\}$ \\ 
\rowcolor{LightCyan}
~ & {\tiny (AUC)} & {\tiny (AUC)} & {\tiny (AUC)} & {\tiny (AUC)} & {\tiny (AUC)} & {\tiny (AUC)} & {\tiny (AUC)} & {\tiny (AUC)} & {\tiny (AUC)} \\ 
\hline\\[-1ex]
Subject 1 & 0.7781 & \textbf{0.8101} & 0.1824 & \textit{0.6987} & 0.7754 & 0.1624 & 0.7141 & \textit{0.6998} & 0.2811 \\ 
\hline\\[-1ex] 
\rowcolor{Gray}
Subject 2 & \textbf{0.8115} & \textit{0.6755} & 0.3045 & 0.7375 & \textit{0.6589} & 0.1847 & 0.7157 & \textbf{0.8095} & 0.1450 \\ 
\hline\\[-1ex] 
Subject 3 & 0.7801 & 0.7787 & 0.2104 & 0.7584 & \textbf{0.8201} & 0.1279 & \textit{0.6590} & 0.7189 & 0.2515 \\ 
\hline\\[-1ex] 
\rowcolor{Gray}
Subject 4 & 0.7124 & \textbf{0.8441} & 0.1980 & \textbf{0.8014} & 0.7352 & 0.1848 & 0.7684 & 0.7871 & 0.2801 \\ 
\hline\\[-1ex]  
Subject 5 & \textbf{0.8380} & 0.7600 & 0.1709 & 0.7412 & 0.7278 & 0.2812 & \textit{0.6971} & 0.7358 & 0.2103 \\ 
\hline\\[-1ex]  
\rowcolor{Gray}
Subject 6 & 0.7312 & 0.7112 & 0.1782 & 0.7813 & 0.7177 & 0.2104 & 0.7300 & 0.7784 & 0.2081 \\ 
\hline\\[-1ex] 
Subject 7 & \textit{0.6819} & 0.7211 & 0.1541 & 0.7987 & 0.\textbf{8380} & 0.1784 & 0.7630 & \textit{0.6798} & 0.2765 \\ 
\hline\hline\\[-1ex] 
\rowcolor{LightCyan}

\textbf{Mean}  & 0.7618 & 0.7572 & 0.1997  & 0.7596 &  0.7533  & 0.1899 & 0.7210 & 0.7441 & 0.2360 \\ 
\hline\\[-1ex] 
\textbf{S.D.} & 0.0557 & 0.0590 & 0.0496  & 0.0371 & 0.0622   & 0.0475 & 0.0378 & 0.0484 & 0.0509 \\ 
\hline\hline\\[-5pt]  
\end{tabular}
\label{table:TABLE 1:Experiment 1} 
\end{table*}

\section{Online Test Results and Analysis}
\label{sec:resultsonline}

In this experiment, 
NIRS reads input brain signals from the subjects
to move the haptic device in real-time.
As in the offline case, we use both MI and AOMI task commands.  
Communication between NIRS and the haptic device is established through the user datagram protocol. 

\begin{figure}[t]
\centering
\includegraphics[width=0.48\textwidth]{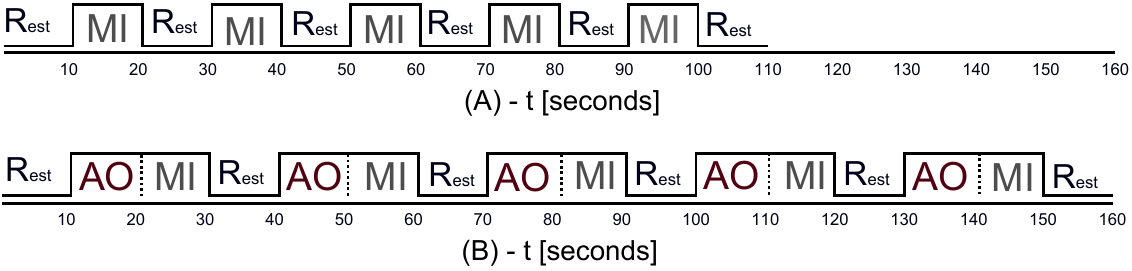}
\caption{Online experimental protocol for data acquisition. (A) Pure motor imagery (MI) task command. (B) Action observation-motor imagery (AOMI) task command. In both experiments, the classifiers decode a user's intent every MI execution.}
\label{fig:protocol}
\end{figure}

The online experimental steps are listed starting from Line 12 onwards of Algorithm~\ref{alg:msvm}. The input data defined in Line 12 correspond to streaming data given in real-time to the optimized classifiers. The final output command is attained on the basis of the AUC value.
We conducted both experiments in at least five sessions, not exceeding one session per day or eight sessions in total for a given subject. 
The sessions were organized by inserting AO from the data acquisition protocol, shown in Fig.~\ref{fig:session}B. We set the timing of session blocks as shown in Fig.~\ref{fig:protocol}. 
The input consisted of test points 3-5$~sec$ long for $\mathcal{E}_1$ and 
$\mathcal{E}_2$, which were equivalent to 42 samples and 20 channels, that is, ${X}\in R^{42\times 20}$. 
The classifier performance is measured during MI task execution periods as shown in Fig.~\ref{fig:protocol}. 

\subsubsection{Experiment 1 (MI Task)}
Table 2 lists classification results from three different sessions corresponding to the pure MI task command. 
In general, lower classification accuracies were obtained in 
the online experiment than in the offline experiments. 
A strong variability is observed in the performances of classifiers across different subjects, sessions, and tasks. 
The mental tasks were more recognizable 
in some subjects than in other subjects, 
resulting in larger deviations in AUC values. 
Classifier performance in the $\Theta_{accept}$ are shown in boldface,
while $\Theta_{worst}$ are italicised.
The trained classifier was successful in    
55 out of the total 63 cases with 
eight unacceptable
cases.


On average, the classifier performances were equivalent to ($AUC = 0.74\pm 0.2$) within the $\Theta_{accept}$ region. We obtained the maximum possible accuracy ($AUC=0.8441$) in classifying  $\{t_{left}\}$ data from Subject 4. 
It is noted that the case of $\Theta_{best}$ performances were not consistent when classifying the same mental task by the same subject. 
This emphasizes the major BCI problem of inter-subject and inter-session variabilities with large standard deviations as shown in the table. 
Even if a participant performed well in one session, the performance within the session may have varied greatly among the $\Theta_{best}$, $\Theta_{accept}$ and $\Theta_{worse}$ regions. 
 
Let us consider the rest task $\{t_{rest}\}$ in all three sessions. We note that during the first 3~$s$ of a task period the classifiers produced increased false positive rates by detecting task-relevant signals as baseline signals. This is because of the high inherent latency of the brain hemodynamic response, which occurs over the interval 4-8~$s$ after the task onset \cite{coffey2010},\cite{gratton2005}. Moreover, we have observed the occurrence of the $U$ case in Eq. 4 when multiple classifiers did not detect any of the mental activity.

\begin{table*}[ht]
\caption{Experiment 2- The online performance of classifiers in decoding signals corresponding to action observation-motor imagery (AOMI) tasks commands of
$\{t_{right}\}$, $\{t_{left}\}$ and $\{t_{rest}\}$.}
\centering 
\begin{tabular}{l| c c c| c c c| c c c} 
\hline\hline\\[-5pt] 
Subjects &~ &  Session 1 & ~ & ~ & Session 2 & ~ & ~ &  Session 3  & ~\\ 
\\[-2ex]
\rowcolor{LightCyan}
~ & $\{t_{right}\}$ & $\{t_{left}\}$ & $\{t_{rest}\}$ & $\{t_{right}\}$ & $\{t_{left}\}$ & $\{t_{rest}\}$ & \{$t_{right}\}$ & $\{t_{left}\}$ & $\{t_{rest}\}$ \\ 
\rowcolor{LightCyan}
~ & {\tiny (AUC)} & {\tiny (AUC)} & {\tiny (AUC)} & {\tiny (AUC)} & {\tiny (AUC)} & {\tiny (AUC)} & {\tiny (AUC)} & {\tiny (AUC)} & {\tiny (AUC)} \\ 

\hline\\[-1ex]
Subject 1 &\textbf{ 0.8103} & 0.7811 & 0.1441 & 0.7217 & \textbf{0.8125} & 0.1341 & \textbf{0.8974} & 0.7489 & 0.1481 \\ 
\hline\\[-1ex] 
\rowcolor{Gray}
Subject 2 & 0.7357 &\textit{0.6982} & 0.2105 & 0.7875 & 0.7510 & 0.1671 &\textbf{0.8300} & 0.7712 & 0.1901 \\ 
\hline\\[-1ex] 
Subject 3 & 0.7508 & 0.7517 & 0.2100 & 0.8780 & 0.7982 & 0.1569 & 0.7124 & 0.7680 & 0.2074 \\ 
\hline\\[-1ex] 
\rowcolor{Gray}
Subject 4 & \textbf{0.8142} & 0.7802 & 0.1870 & 0.7984 & \textit{0.6815} & 0.1908 & \textbf{0.8670} & 0.7046 & 0.2011 \\ 
\hline\\[-1ex]  
Subject 5 & 0.7918 & \textit{0.6901} & 0.2650 & 0.7550 & 0.7201 & 0.2233 & 0.7919 & \textbf{0.9308} & 0.1030 \\ 
\hline\\[-1ex]  
\rowcolor{Gray}
Subject 6 & \textbf{0.8301} & 0.7011 & 0.1982 & \textbf{0.8183} & 0.7870 & 0.1789 & 0.7710 & 0.7118 & 0.1900 \\ 
\hline\\[-1ex] 
Subject 7 & 0.7416 & 0.7809 & 0.1455 & 0.7004 & 0.7909 & 0.1399 & 0.7321 & 0.7808 & 0.1987\\
\hline\hline\\[-1ex] 
\rowcolor{LightCyan}
\textbf{Mean} & 0.7820 & 0.7407 & 0.1923 & 0.7799 & 0.7630 & 0.1701 & 0.8002 & 0.7737 & 0.1769 \\ 
\hline\\[-1ex] 
\textbf{S.D.} & 0.0387 & 0.04255 & 0.0417 & 0.0603 & 0.0477 & 0.0309 & 0.0683 & 0.0752 & 0.0379 \\ 
\hline\hline\\[-5pt]  

\end{tabular}
\label{table:TABLE 1} 
\end{table*}

\subsubsection{Experiment 2 (AOMI Task)}
Table 3 lists the decoding results of signals
corresponding to AOMI task commands. Compared to 
the MI task commands, AOMI task commands achieved superior accuracy.
AOMI task 
commands were successful in
60 out of the total 63 cases with only three
cases of failure.

Let us compare some specific results between MI and AOMI experiments.
For instance, consider the Table 2 entries for Subject 1 during Session 2  $\{t_{right}\}$, and Session 3 $\{t_{left}\}$ when the classifier accuracies were in the  $\Theta_{worst}$ region. In contrast, the corresponding entries of Table 3 show much improvement in the AUC values from the AOMI experiments. 

By 
comparative analysis, we conclude the following.
First,
the average AUC values in the AOMI experiment were not significantly better than those of pure MI tasks. However, individual comparisons show improvements by subjects between sessions.
Second, the standard deviations of AUC values were in a range similar to that of the last experiment. And lastly,
the dominating inter-subject and inter-session variabilities were  observed in terms of AUC values on both MI and AOMI experiments.

\begin{figure*}
\centering
\includegraphics[width=0.99\textwidth]{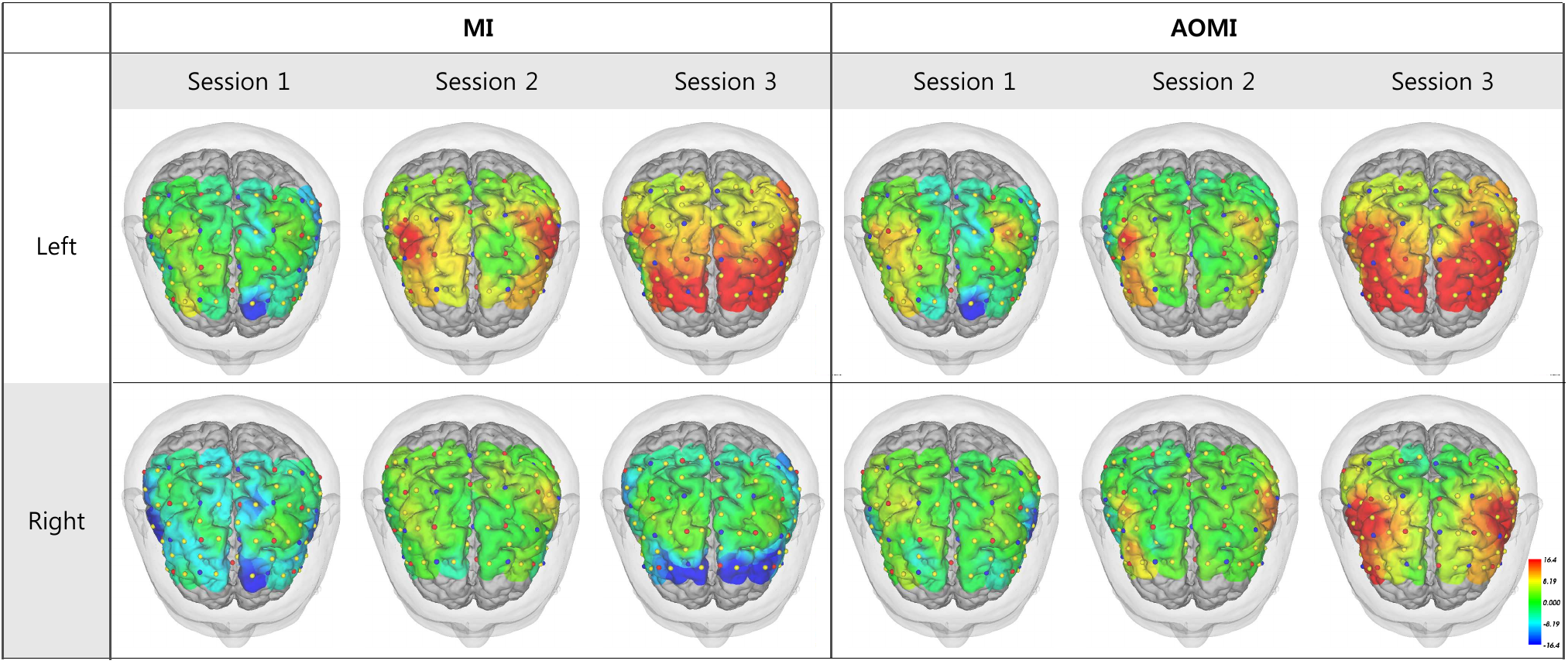}
\caption{Representative cortical mapping of oxy-Hb level changes related to MI and AOMI tasks. The data were obtained from Subject 1 over three sessions. The color scale indicates the coordinates of concentration changes in oxy-Hb with t-values.}
\label{fig:mapping}
\end{figure*}

\subsubsection{Brain Mapping Analysis}

We assumed that the classifier performances were affected by the variability in the task-relevant cortical activation areas. Because the exact locations of the task-relevant channels were not always the same, we further analyzed a topographic cortical mapping of task-relevant oxy-Hb level changes by using a general linear model (GLM) algorithm explained in \cite{schroeter2004}. The significance thresholds for the statistical parametric maps were set to 
$p<0.05$.

We separated the topographic map into nine regions of interest according to the functional anatomy of the premotor and prefrontal regions including the sensorimotor cortex (SMC), supplementary motor area (SMA), presupplementary motor area (preSMA), dorsal premotor cortex (PMC), and dorsolateral prefrontal cortex (PFC). The right lateral SMC was covered by Channels 1, 7, 8, and 9; the left lateral SMC by Channels 5, 6, 11, 12, and 13; the SMA by Channels 16, 17, 22, and 23; the preSMA by Channels 35, 36, 42, and 43; the right PMC by Channels 14, 15, 20, and 28; the left PMC by Channels 18, 19, 24, 25, and 26; the left PFC by Channels 27, 33, 34, 40, and 41; and the right PFC by Channels 31, 32, 37, 38, 39, 44, and 45.

\begin{table*}[ht]
\caption{Analysis methods used in previous NIRS-BCI studies. The asterisks denote online classification results} 
\centering 
\begin{tabular}{l| l| l| l| l} 
\hline\hline\\[-5pt] 
\\[-2ex]
Author (Ref) & Brain region & Input features & Classifier & Performance \\ 
\hline\\[-1ex]

\textbf{Our study}  & Prefrontal, & OxyHb & Ensemble SVM &\textbf{76\%}-\textbf{93\%}$^{*}$ \\ 
~ & Sensorimotor  & PCA & Classifiers & ~ \\ 
~ & cortex &  &  & ~ \\ 
\hline \\[-1ex]

Abdelnour (\cite{abdelnour2009}) & Motor cortex & OxyHb after & Linear discriminant & 68\%-100\%$^{*}$  \\ 
~ & ~ & Kalman filtering &  analysis (LDA)  & ~ \\ 
\hline\\[-1ex]

Coyle (\cite{coyle2004b}) & Motor cortex & mean OxyHb & OxyHb amplitude & 80\%$^{*}$ \\ 
~ & ~ &  & threshold detector & ~ \\ 
\hline \\[-1ex]

Utsugi (\cite{utsugi2007}) & Prefrontal & OxyHb & Artificial Neural & 70\%-90\%$^{*}$ \\ 
~ & cortex & ~ & Networks & ~ \\ 
\hline\\[-1ex]

Abibullaev (\cite{abibullaev2012}) & Prefrontal & OxyHb wavelet & Linear discriminant & LDA 81\%-95\%,\\ 
~ & cortex & coefficients & analysis , Artificial &  ANN 69\%-91\% \\ 
~ & ~ & ~ &  Neural Networks, SVM & SVM 94\%-97\%  \\ 
\hline\\[-1ex]

Coyle (\cite{coyle2004a}) & Motor cortex & mean OxyHb & Simple threshold & 75\% \\ 
~ & ~ & of 20sec data &  detector & ~ \\ 
\hline\\[-1ex]

Cui (\cite{cui2010}) & Motor cortex & different features & Support Vector & 70\% - 90\% \\ 
~ & ~ & OxyHb, deOxyHb & Machines & ~ \\ 
\hline\\[-1ex]

Fazli (\cite{fazli2012}) &  Prefrontal, & EEG\& fNIRS  & Linear Discriminant & 78.6\%-92.9\% \\ 
~ & Motor cortex & Hybrid features &  Analysis & ~ \\ 
\hline\\[-1ex]

Sassaroli (\cite{sassaroli2008}) & Prefrontal & OxyHb, deOxyHb & K-means & 55.6\%-72.2\% \\ 
~ & cortex & raw features & algorithm & ~ \\ 
\hline\\[-1ex]

Sitaram (\cite{sitaram2007}) & Frontal & OxyHb intensity  & Hidden Markov & SVM 73\% \\ 
~ & cortex & ~ & Models (HMM), SVM & HMM 89\% \\ 
\hline\\[-1ex]

Tai (\cite{tai2009}) & Prefrontal & OxyHb intensity  & LDA, SVM & 75\%-96\% \\ 
~ & cortex & ~ & ~ & ~ \\ 
\hline\\[-1ex]

Truong (\cite{truong2009}) & Prefrontal & OxyHb wavelet & Artificial Neural & 95\% \\ 
~ & cortex & decomposition & Networks (ANN) & ~ \\ 
\hline \hline

\end{tabular}
\label{table:TABLE 1} 
\end{table*}

The location of a Cz reference point is represented by Channel 10 (see Fig. 2(A)).
Fig. 5 shows a distinct cortical activation pattern reconstructed from data on Subject 1 across different sessions for both MI and AOMI tasks. Task-relevant increases of oxy-Hb were prominent in the prefrontal regions but were strongly dependent on the task type and the session type. For instance, with repetition of the session the increase of oxy-Hb appeared to intensify for the channels covering the right and left PMC for both MI and AOMI while performing the $\{t_{left}\}$ task. The oxy-Hb was augmented in the channels covering the SMA and remained unchanged in the channels covering the left SMC. For the MI-based $t_{right}$  task, we have observed the reverse case; that is, with repetition of the session the oxy-Hb concentration levels were observed to decrease in the pre-SMA and PFC regions. By using Algorithm\ref{alg:rce} each time, we tend to select for a classifier only those task-relevant channels with higher activations. Therefore, each time the locations of task-relevant channels vary, the performance of a classifier is affected. In general, cortical activation in the pre-SMA and PFC remained relatively unchanged among most of the subjects within a session. We visually inspected all changes in the regional activation with respect to the subject, task, and session type.

We briefly summarize our findings of the mapping analysis as follows:

\begin{itemize}
\item A session-dependent cortical activation was seen for both MI and AOMI tasks.
\item In some subjects, the cortical activation levels increased with the number of sessions.  
\item The AOMI task produced higher cortical activation than the MI task. The major activation locations for the AOMI task included the PFC, PMC, SMA, and pre-SMC regions. 
\item The effect size calculated by using oxy-Hb levels showed no significant difference in either the $\{t_{right}\}$ $(p = 0.203)$ or the  $\{t_{left}\}$ $(p = 0.535)$. In terms of the course of oxy-Hb changes during the AOMI period, two tasks showed comparable intervals between the start of the $\{t_{right}\}$ task and the peak of oxy-Hb in the $\{t_{left}\}$.
\item No strong correlation between the MI and AOMI tasks was observed. The effect of preparation on the increases in oxy-Hb level during the MI task and AOMI task was evaluated by calculating effect sizes. In terms of oxy-Hb levels, a one-way ANOVA showed a significant main effect for site during both the MI period $(p < 0.05)$ and in the AOMI period $(p < 0.05)$. 
\end{itemize}

\section{Discussion}
\label{sec:discussion}

We have shown that it is possible to command
a haptic device to move in opposing directions 
by detecting oxy-Hb signal reading from NIRS-BCI system.
This study proposes that such capability can be used
for neurorehabilitation to induce brain plasticity
for stroke patients, or possibly provide them with
some degree of self-sufficiency.
MI and AOMI
task commands were implemented in both
online and offline modes. 
Feature extraction and channel localization reduced
noise in the input signals for classification of multiple
SVMs.
The online BCI classification of pure motor imagery tasks was 76\% accurate on average. And we observed a significant improvement in the BCI accuracies of up to 93\% when using signals from 
AOMI task compared to pure MI.
Compared to other studies, ours has obtained improved classification rates, as shown in Table 4. 
Note that only a few online classification results
achieved performances in the range of 70\%-90\% \cite{coyle2007}, \cite{utsugi2007}. Except for the results of Abdelnour et al. \cite{abdelnour2009} whose online classification rate ranges from 68.8\% to 100\%,
our work showed better results.
However, it is noted that
\cite{abdelnour2009} used real finger tapping 
which is more discriminable than pure mental task.

The methodology proposed in this paper differs from that in the other studies by virtue of the following attributes:

\begin{itemize}
\item We use 45-channels recordings which cover the most important regions of the brain cortex (SMC, SMA, and PFC). This is in contrast with other NIRS-BCI methods, which usually cover minimal locations of the brain cortex . We then perform an automated channel selection method which allowed us to localize 20 most task relevant channels for subsequent classification. Using multiple channels can be seen as a disadvantage, in general.  However, our motivation was to accurately localize task relevant channels each time when subjects perform a specific mental task. Further, because we conduct two different experiments with new tasks (the imaginary directional movement tasks are not common BCI research) there was a need to investigate multiple channel recordings. Moreover, before each online experiment we perform a channel localization within minutes in each session and for each subject. Such localization helps to capture the session or subject specific neural activation within the few relevant channels. For instance, we have noticed that the classifier uses different channel combinations to discriminate between the imaginary rightward movement and the imaginary leftward movement. In case the fixed number of channels over a specific brain region were used then the classification accuracy could be very unsatisfactory. This is a different mechanism which allowed us to automatically switch the strong task relevant channels among various mental tasks. 

\item We performed different BCI experiments that included directional hand movement tasks based on pure motor imagery (MI) tasks and combined action observation and motor imagery (AOMI) tasks. We found that AOMI tasks were more classifiable compared to the pure MI tasks. Our idea of designing the directional movement tasks are intended for application in stroke rehabilitation physical therapy, which is envisaged to combine BCI with therapeutic devices for upper limb exercises. We have extracted the tasks after reviewing the important tasks used in stroke rehabilitation to improve the activities of daily living (ADL). The example of other tasks include "reaching", "pulling", "flexion" and "extension" of upper limb. Among them in our initial phase we implemented ``leftward'' and ``rightward'' movements. The haptic device was just a test platform, however it can be easily replaced with the available upper limb physical therapy devices (e.g. MIT Manus). The limitation of our study is that we study only normal subjects at present study. Moreover, the number of subjects are limited to seven. Because our initial goal was to verify the feasibility of our BCI approach. Nonetheless, we measured the data from extensive number of sessions to support the potential of the present approach. Another important question is that whether the AOMI task is effective for effective neuro-rehabilitation (e.g. improved cortical re-organization or neuroplasticity). We focus to study these question in our future study. Due to the possible application of our study, we put higher priority in the detection of MI tasks in a direct (non-interpretive) way to provide natural BCI outputs. 
\item We presented a different classification approach which is robust against the major BCI classification problems. Because we have optimized the classifiers in the offline settings from vast data from as many sessions as possible, they performed robust throughout online experiments. There were only few exceptions of lower classification results.  
To date, most NIRS-BCI studies used standard classifiers such as SVMs, LDA, HMM, or ANNs from Table 3. This study presents another classifier which achieves higher accuracies by using a multiple learning strategy. However, it is not appropriate to compare and judge the research results in terms of classification accuracy, because many factors influence the difficulty and the accuracy in a particular BCI study. For instance, such factors include the type of BCI paradigm and whether the NIRS signal characteristics used are raw, preprocessed, or transformed. In addition, the type of the system used to acquire the NIRS signals is a factor. In our study, the signal sampling frequency was 14.28 Hz, whereas other NIRS-BCIs use signals with sampling rates from 2 Hz to 10 Hz. The lower the sampling frequency, the lower the signal quality and harder the extraction of the true neural signals from background noise. 
\end{itemize}

One known limitation of the present NIRS-BCI approach is the delay in operating a haptic device because of the intrinsic latency of the brain hemodynamics. With an EEG-BCI system, an operation can be performed over a few milliseconds. 
We plan to experiment two possible ways of overcoming the slowness of the NIRS-BCI that we aim to research in the next step. The first is based on exploring fast hemodynamic responses as was done by Cui et al. (2010) . The other is to develop a hybrid BCI paradigm that combines EEG and NIRS signals for rapid detection of mental state as in (Fazli et al., 2012). In addition, we plan to extensively study the influence of the various feedback types (visual, auditory, or haptic) and their effects on the improvement of the overall classification accuracies. In general, the NIRS-BCI may not be suitable for a fast translation of mental intent, however we believe that it has potential for a neurorehabilitation and motor learning of post-stroke patients that involves slow operations.

\end{document}